\documentclass{article}
\usepackage{spconf,amsmath,graphicx}

\usepackage{caption}
\usepackage{amsmath,epsfig}
\usepackage{subfigure}
\usepackage{amscd}
\usepackage{color, colortbl}
\usepackage{amssymb}
\usepackage{bm}
\usepackage{bbm}

\usepackage{algorithm}
\usepackage{algpseudocode}
\usepackage{balance}
\usepackage{booktabs}
\usepackage{multirow}

\usepackage{xcolor}

\title{ A conformalized Learning of a Prediction Set with Applications to Medical Imaging Classification }

  \name{ Roy Hirsch \qquad Jacob Goldberger  }
\address{  Faculty of Engineering, Bar-Ilan University, Ramat-Gan, Israel
\\
\texttt{jacob.goldberger@biu.ac.il}
\thanks{This research was supported by the Ministry of Science \&  Technology, Israel.}
}

\begin{document}
\maketitle

\begin{abstract}
Medical imaging classifiers can achieve high predictive accuracy, but quantifying their uncertainty remains an unresolved challenge, which prevents their deployment in medical clinics. We present an algorithm that can modify any classifier to produce a prediction set containing the true label with a user-specified probability, such as 90\%. 
We train a network to predict an instance-based 
 version of the Conformal Prediction threshold. The threshold is then conformalized to ensure the required coverage.
We applied the proposed algorithm to several standard medical imaging classification datasets. The experimental results demonstrate that our method outperforms current  approaches in terms of smaller average size of the prediction set while maintaining the desired 
coverage.

	\end{abstract}

\section{introduction}

Deep learning holds immense potential for automating numerous clinical tasks in medical imaging. However, the translation of black-box deep learning procedures into clinical practice has been impeded by the absence of transparency and interpretability. Many clinical deep learning models lack rigorous,  robust techniques for conveying their confidence in their predictions, which limits their appeal for widespread use in medical decision-making. In the case of  medical imaging classification tasks, in addition to the  most likely diagnosis, it is equally or more important to rule out options. Decision reporting in terms of a prediction set of class candidates is thus a natural approach for clinical applications. In this procedure, most classes are ruled out by the network, leaving the physician with only a few options to further investigate. In terms of confidence, we expect that the prediction set provably covers the true diagnosis with a high probability (e.g., 90\%).

Conformal Prediction (CP) \cite{vovk2005conformal, angelopoulos2021gentle} is a general non-parametric calibration method  that was initially formulated for the task of classification, and was later modified for regression.
In the case of classification, given a confidence level
$1\!-\!\alpha$, it aims to build a small prediction set with a guarantee that the probability that the correct class is within this set is at least  $1\!-\!\alpha$.
CP was invented more than twenty years ago, before deep learning have emerged. Recently, it has become a major calibration tool for neural network systems in various applications including  medical imaging \cite{lu2022fair,lu2022improving}. CP is not a specific algorithm but rather a general framework in which selecting a specific conformal score (aka non-conformity score) defines the way the prediction set is constructed. The parameters of the CP algorithm are tuned on a validation set to ensure the required coverage of the prediction set. To make CP effective, the size of the prediction set should be as small as possible. The effectiveness of different CP variants is thus measured by calculating the average size of the prediction sets on the test set. 

Note that in addition to the CP calibration algorithm, which focuses on forming a prediction set with guaranteed coverage, there is another family of calibration algorithms whose goal is to tune the confidence of the predicted class. The most widely used strategy is post-hoc calibration of the softmax logit 
scores, e.g., Temperature Scaling  \cite{Guo2017, nixon2019measuring}. Here, the calibration goal is different in that it involves forming a small but reliable prediction set.

The Adaptive Prediction Sets (APS) score, which was first introduced in  \cite{romano2020classification}, is a commonly used conformal score.  The APS algorithm generates a prediction set by selecting the most likely classes until the cumulative probability exceeds a predetermined threshold. The Regularized Adaptive Prediction Sets (RAPS) algorithm 
\cite{angelopoulos2020uncertainty}
is a variant of the APS method that modifies its conformal score by penalizing prediction sets that are too large.
%
%
{This approach is particularly useful for classification tasks with many possible labels. It returns predictive sets that achieve a pre-specified error level while retaining a small average set size.}
All current CP variants, including APS and RAPS, are based on finding a single threshold, denoted as $q$, such that the conformal scores of most of the validation-set samples are below $q$. This implies that for some of the samples, $q$ is overly pessimistic, leading to unnecessarily large prediction sets.

In this study, we introduce a CP algorithm in which the threshold used to create the prediction set is optimized for each sample individually. The threshold is computed by a neural network and then adjusted (conformalized) to ensure that it meets the coverage requirements. We applied the proposed algorithm to several standard medical imaging classification datasets. The experimental results demonstrate that our method outperforms existing state-of-the-art CP methods significantly, in terms of the average size of the prediction set while maintaining the same coverage.

\section{Conformal Prediction}
In this section we  review the commonly used variants of the Conformal Prediction  algorithm \cite{vovk2005conformal} and  in the next section we present our approach within this framework.

Consider a network that classifies an input $x$ into $k$ pre-defined classes.
Given an input $x$ with an unknown class $y$ we want to find a small prediction set of classes $C(x)\subset\{1,...,k\}$ such that $p(y\in C(x)) > 1\!-\!\alpha$
where $\alpha$ is a predefined miss-coverage rate.
A simple approach to achieving  this goal is to include classes from the highest to the lowest probability 
 until their sum just exceeds the threshold $1\!-\!\alpha$. As in \cite{angelopoulos2020uncertainty},  this uncalibrated prediction-set strategy is dubbed here `naive'. 
 {While the network output has the mathematical form of a distribution, this does not necessarily imply that it represents the true class distribution.} The network is usually not calibrated and tends to be over-optimistic \cite{Guo2017}. 

The CP algorithm builds a prediction set (with a probabilistic justification) in the following way. Let $x$ be a sample along with its network-based class probabilities $p_1,...,p_k$.
Let $\pi(x)$ be the permutation of $\{1,...,k\}$  sorted  from the most likely class to the least likely, i.e.
$$p(\pi_{1}) \ge \dots  \ge p(\pi_{k}).$$
For each $v\in [0,1]$ we form the prediction set $C_{v}(x)$ by taking top-scoring classes until the total mass just exceeds $v$.
More formally,  $C_{v}(x) = \{\pi_1(x), \dots, \pi_l(x)\}$ such that:
\begin{equation} l =  \min \{ l' | \sum_{i=1}^{l'} p(\pi_i)  \ge v \}. 
\label{apsindex}
\end{equation}
Given a labeled sample $(x,y)$ s.t. 
$y\in\{1,\dots,k\}$, the Adaptive Prediction Score (APS) \cite{romano2020classification} is defined to be the set of all the classes whose score is greater or equal to the score of the true label $y$:
\begin{equation}
 s(x,y) =  \sum_{\{i|p_i \ge p_{y}\}} p_i
  \label{aps_score}
  \end{equation}
The score $s(x,y)$ is the minimal $v\in[0,1]$ in which the true class $y$ is in a prediction set  $C_v(x)$.
We can define a randomized version of 
(\ref{aps_score})  as follows: 
\begin{equation}
 s_{\textrm{random}}(x,y) =  u \cdot p_y + \sum_{\{i|p_i > p_{y}\}} p_i  
  \label{aps_score_rand}
  \end{equation}
s.t. $u$ is a r.v. uniformly distributed in the interval $[0,1]$.
The randomization 
can help to achieve $1\!-\!\alpha$ coverage exactly on the validation set 
\cite{einbinder2022training}.
The {Regularized Adaptive
Prediction Sets (RAPS)} score \cite{angelopoulos2020uncertainty} is a variant of APS that encourages small prediction sets. It is defined as follows:
\begin{equation}
 s(x,y) =  \sum_{\{i|p_i \ge p_{y}\}} (p_i +
 a \cdot 1_{\{i>b\}})
   \label{raps_score}
  \end{equation}
s.t. $a$ and $b$ are parameters that needed to be tuned.

Let $(x_1,y_1), \dots, (x_n,y_n)$ be  a labeled 
validation set and let $s_t=s(x_t,y_t)$ be the 
conformal score of $(x_t,y_t)$.
$s_t$ is the minimal threshold in which the true class $y_t$ is in a prediction set of $x_t$.
Let $q$ be the $(1\!-\!\alpha)$ quantile of $s_1,...,s_n$,
i.e. if we sort the values $s_1 \le \dots \le  s_n$ then 
$q=s_{(1\!-\!\alpha)n}$. 
The threshold $q$ is the minimal one in which  the correct label $y_t$ is included in the prediction set $C_{q}(x_t)$ for at least  $(1\!-\!\alpha)n$ points of the validation set. Given a new test point $x$, we report the prediction set $C_{q}(x)$. 
The general CP theory \cite{vovk2005conformal} guarantees that: $$1\!-\!\alpha \le p( y\in C_q(x)) \le (1\!-\!\alpha) + \frac{1}{n+1}, $$
where $y$ is the unknown true label.
Note that this is a marginal probability over all
possible  test points and is not conditioned on a given input.


   

\begin{algorithm}[t]
\caption{Conformalized Prediction Set Network (CPSN)}

    \vspace{1mm}
    \textit{Training Phase:}
    \vspace{1mm}
    
    \begin{algorithmic}[1]
    \State Given a training set $(x_1,y_1),\dots,(x_n,y_n)$,
    calculate the APS conformal score per instance $s(x_t,y_t)$.
    \State Learn a network $q(x;\theta)$ by minimizing the MSE loss: 
    $$
     l(\theta) = \sum_t  (q(x_t;\theta)-s(x_t,y_t))^2.
    $$    
    \end{algorithmic}
    
    \hrule
    \vspace{1mm}
    \textit{Conformalization Phase:}
    \vspace{1mm}
    
    \begin{algorithmic}[1]
    \State Given a validation set $(x_1,y_1),...,(x_n,y_n)$,
    predict the conformal score per instance $q(x_t;\theta)$.
    \State Define 
    $$
    \begin{aligned}
    \mathbb{G}_1 = \{r_t | \hat{p}(x_t) > 1\! - \!\alpha\},  \,\,\,
    \mathbb{G}_2 = \{r_t | \hat{p}(x_t) \leq 1\! - \!\alpha\}
    \end{aligned}
      $$
    s.t. $r_t = s(x_t,y_t) - q(x_t)$ and  $\hat{p}(x_t) \!=\! \max_ip(y_t\!=\!i|x_t)$.
    \State 
Define: 
    $
    \delta_i =  \lceil(n\!+\!1)(1\!-\!\alpha)/n\rceil \text{-quantile of } \mathbb{G}_i
    , \,\,\, i=1,2$
    
    \end{algorithmic}
    
    \vspace{1mm}
    \hrule 
    \vspace{1mm}
    {Usage:} Given a test sample $x$:
    \begin{algorithmic}[1]
    \State If $\hat{p}(x)>1\!-\!\alpha$ define $\delta(x)=\delta_1$ and otherwise $\delta(x)=\delta_2$.
    \State Report the prediction set $C_{(q(x)+\delta(x))}(x)$. 
     \end{algorithmic}
    \vspace{1mm}
    \hrule
    \vspace{1mm}
     There is a guarantee:
$p( y\in C_{(q(x)+\delta(x))}(x)) \ge 1-\alpha$.
    \label{algo:main}
    \end{algorithm}

\begin{table*}[]
\label{tab:results}
\small	
    \centering
    \resizebox{\textwidth}{!}{
\begin{tabular}{lcccccccc}
\toprule
 & \multicolumn{4}{c}{\textbf{OrganAMNIST}} & \multicolumn{4}{c}{\textbf{TissuMNIST}} \\
 & \multicolumn{2}{c}{$\alpha=.1$} & \multicolumn{2}{c}{$\alpha=.05$} & \multicolumn{2}{c}{$\alpha=.1$} & \multicolumn{2}{c}{$\alpha=.05$} \\ \hline 
Model & Size $\downarrow$ & Coverage & Size $\downarrow$ & Coverage & Size $\downarrow$ & Coverage & Size $\downarrow$ & Coverage \\ \hline\hline
{Naive} & 2.646 $\pm$ .031 & .950 $\pm$ .003 & 3.444 $\pm$ .054 & .971 $\pm$ .003 & 2.678 $\pm$ .020 & .945 $\pm$ .003 & 3.339 $\pm$ .019 & .973 $\pm$ .002 \\
{APS} & 4.096 $\pm$ .059 & .899 $\pm$ .005 & 5.179 $\pm$ .071 & .952 $\pm$ .004 & 3.464 $\pm$ .031 & .899 $\pm$ .006 & 4.337 $\pm$ .037 & .949 $\pm$ .004 \\
{Rand APS} & 2.289 $\pm$ .030 & .922 $\pm$ .004 & 3.005 $\pm$ .033 & .958 $\pm$ .004 & 2.335 $\pm$ .023 & .912 $\pm$ .005 & 2.956 $\pm$ .023 & .953 $\pm$ .004 \\
{RAPS} & 2.586 $\pm$ .027 & .947 $\pm$ .003 & 3.325 $\pm$ .040 & .968 $\pm$ .003 & 2.676 $\pm$ .021 & .944 $\pm$ .004 & 3.318 $\pm$ .018 & .972 $\pm$ .003 \\
{Rand RAPS} & 2.259 $\pm$ .033 & .900 $\pm$ .006 & 2.977 $\pm$ .035 & .950 $\pm$ .003 & 2.319 $\pm$ .021 & .898 $\pm$ .005 & 2.944 $\pm$ .025 & .948 $\pm$ .003 \\
{CPSN} & \textbf{2.251 $\pm$ .034} & .935 $\pm$ .005 & \textbf{2.818 $\pm$ .073} & .964 $\pm$ .003 & \textbf{2.210 $\pm$ .023} & .906 $\pm$ .003 & \textbf{2.761 $\pm$ .020} & .953 $\pm$ .003 \\ \bottomrule 
    \end{tabular}
        }

\caption{Results on \textbf{OrganAMNIST} and \textbf{TissuMNIST} datasets. We report coverage and size for each model for two $\alpha$ values. We ran each experiment 10 times with different random splits and report the mean and standard deviation. For each setup, the best-performing results are bolded.}

\end{table*}

\section{An Instance-based Threshold for a Prediction Set}

Standard CP algorithms such as APS 
find and apply a single threshold to generate the prediction set for all the  test samples.
We next present an instance-based CP version where {a parametric model is used to predict a different threshold for each sample.} We hypothesize that a per-sample threshold will enable us to generate smaller and therefore more informative prediction sets.

We first train a neural network to predict the APS score $s(x,y)$  (\ref{aps_score})
directly from the feature  input vector $x$, without using the true class $y$. Given a training set $(x_1,y_1),\dots,(x_n,y_n)$,
we learn a regression network by minimizing the following Mean Squared Error (MSE) loss:
\begin{equation}
 l(\theta) = \sum_t  (q(x_t;\theta)-s(x_t,y_t))^2 
\end{equation}
such that $\theta$ is the network parameter set and
$q(x_t;\theta)$ is the network output. 

Since $q(x)$ is learned by a network it is not guaranteed to satisfy the $\alpha$ miss-coverage requirement. We next conformalize the learned threshold to obtain the required coverage.
 Assume we have a labeled validation set $(x_1,y_1),...,(x_n,y_n)$.
 Define the following conformity score {as the residual}:
  \begin{equation}
      r_t = s(x_t,y_t) - q(x_t), \hspace{1cm} t=1,...,n.
  \end{equation}
   For each $t$, $s(x_t,y_t)$ is the minimum value $v$ such that $y_t\in C_{v}(x_t)$. Therefore, $r_t$ is the minimal value (either positive or negative) such that $y_t\in C_{(q(x_t)+r_t)}(x_t)$. {The residual scores act differently for predictions with higher and lower confidence levels. Hence,  we can separate them into two sets based on whether the prediction confidence $\hat{p}(x)=\max_i p(y=i|x)$ is larger or smaller than $1-\alpha$. We calculate the $(1\!-\!\alpha)$ quantile for each of the two sets of scores and define a function $\delta(x)$ that returns the appropriate quantile given the maximal prediction probability. The bias correction function $\delta(x)$ enforces the conformity of the method.

     
Finally, given a test sample $x$ we report the following prediction set: $C_{q(x)+\delta(x)}(x)$.
In other  words, we form the prediction set by 
summing the probabilities of the highly scored classes until we reach $q(x)+\delta(x)$.
  The general CP theory \cite{angelopoulos2021gentle} guarantees that: \begin{equation}
        1\!-\!\alpha \le 
        p( y\in C_{ q(x)+\delta(x)}(x) )
        \le (1\!-\!\alpha) + \frac{1}{n+1},
  \end{equation}
where $y$ is the unknown true label and $n$ is the size of the validation set.
As in other CP algorithms, this is a marginal probability over all possible test points and is not conditioned on a given input. The resulting method, which we call Conformalized Prediction Set Network (CPSN), is summarized in Algorithm Box 1.

\begin{table*}[h!]
    \centering
    \small
    \begin{tabular}{lcccc}
        \toprule
        \multicolumn{3}{c}{\textbf{TissueMNIST}} & \multicolumn{2}{c}{\textbf{OrganAMNIST}} \\ \hline
        \multicolumn{3}{p{85mm}}{Labels = \{Collecting Duct, Connecting Tubule (\textit{C}), Distal Convoluted Tubule (\textit{DCT}), Glomerular Endothelial Cells (\textit{GEC}), Interstitial Endothelial Cells (\textit{IEC}), Leukocytes (\textit{L}), Podocytes (\textit{P}), Proximal Tubule Segments (\textit{PTS}), Thick Ascending Limb (\textit{TAL})\}} & \multicolumn{2}{p{85mm}}{Labels = \{Bladder (\textit{B}), Femur Left (\textit{FL}), Femur Right (\textit{FR}), Heart (\textit{H}), Kidney Left (\textit{KL}), Kidney Right (\textit{KR}), Liver (\textit{L}), Lung Left (\textit{LL}), Lung Right (\textit{LR}), Pancreas (\textit{P}), Spleen (\textit{S})\}} \\ \hline
        & \includegraphics[width=0.07\textwidth]{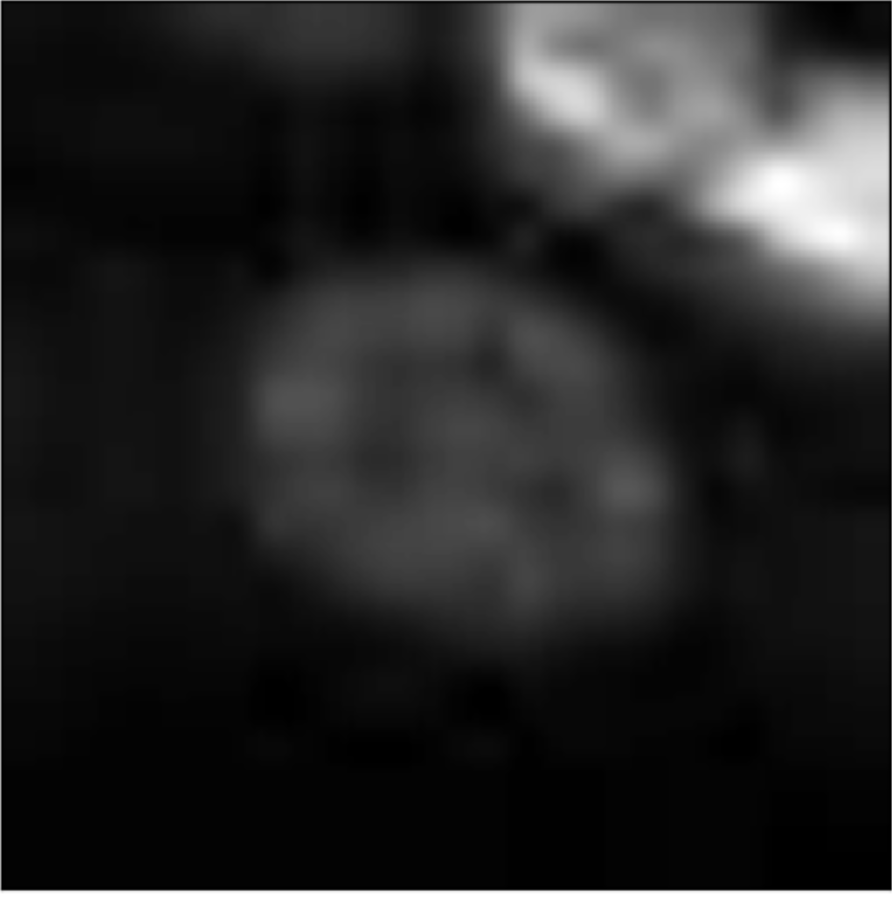} & \includegraphics[width=0.07\textwidth]{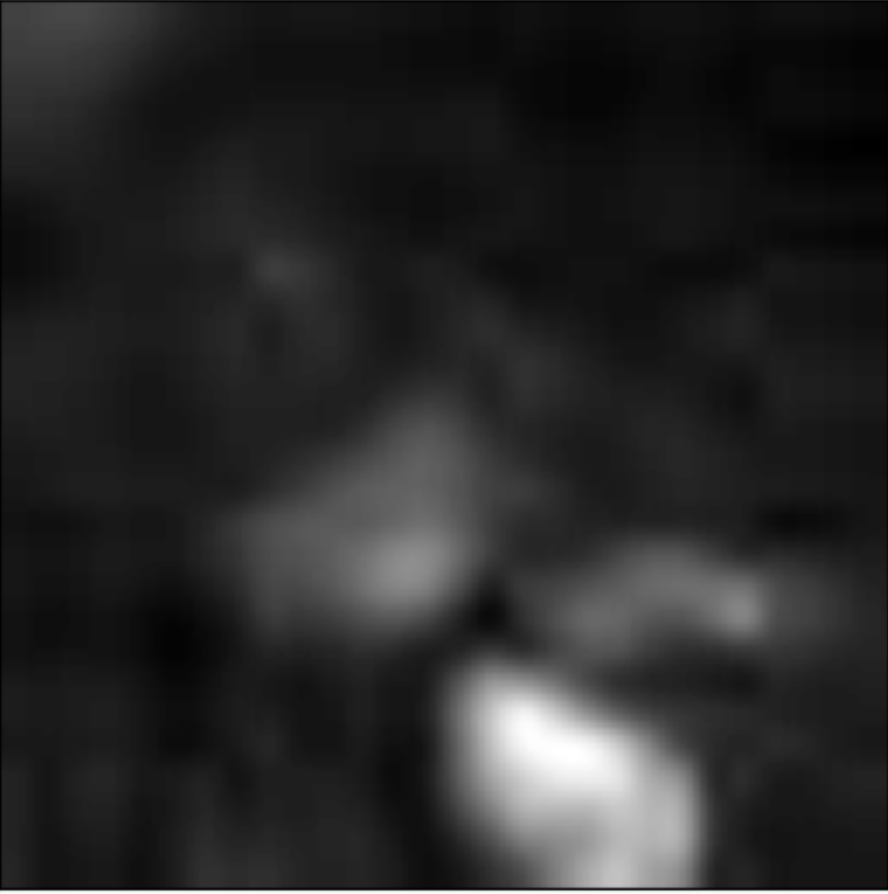} &
        \includegraphics[width=0.07\textwidth]{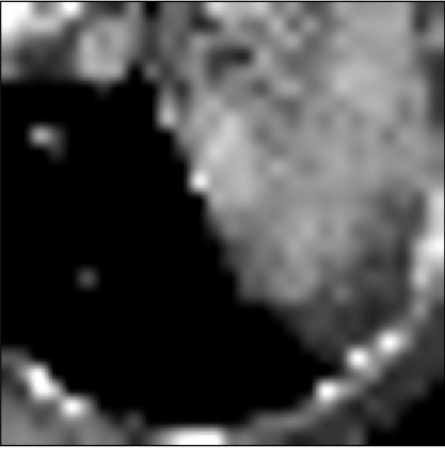} &
        \includegraphics[width=0.07\textwidth]{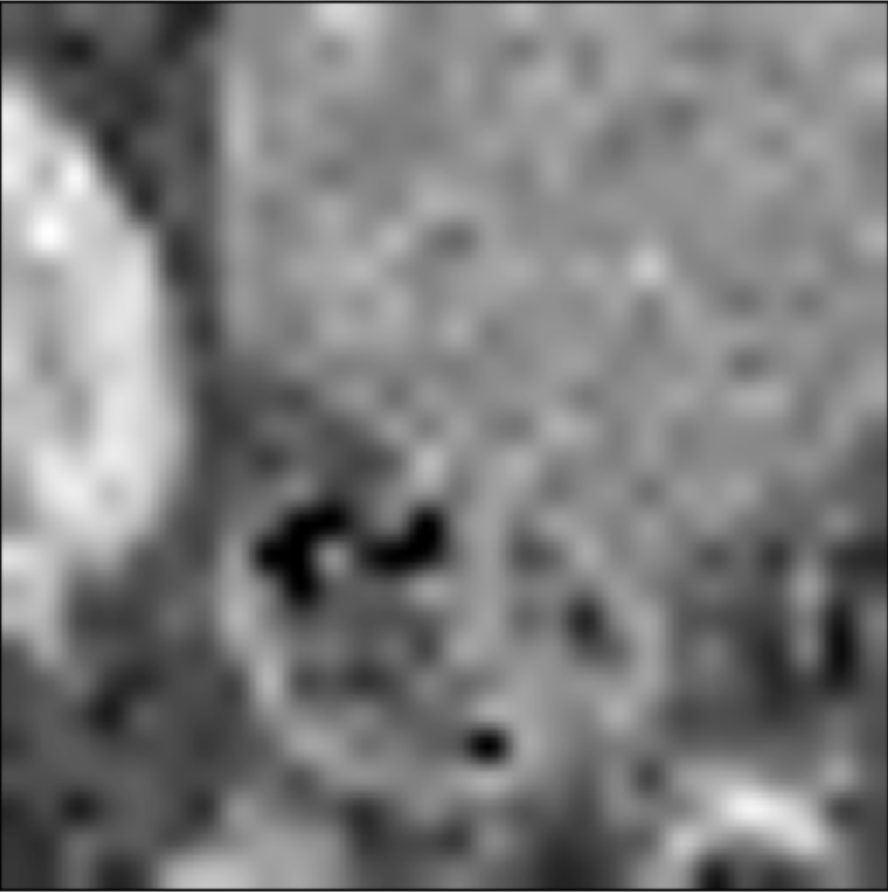} \\ \hline
        \textbf{Naive} & \{C, PTS, \textcolor{green}{TAL}\} & \{IEC, C, PTS, L, P, \textcolor{green}{DCT}\} & \{KL, B, S, \textcolor{green}{LL}\} & \{B, KL, S, \textcolor{green}{KR}\}\\ 
        \textbf{APS} & \{C, PTS, \textcolor{green}{TAL}\} & \{IEC, C, PTS, L, P, \textcolor{green}{DCT}\} & \{H, P, KL, B, S, \textcolor{green}{LL}\} & \{FR, LR, P, B, KL, S, \textcolor{green}{KR}\} \\ 
        \textbf{Rand APS} & \{C, PTS, \textcolor{green}{TAL}\} & \{C, PTS, L, P, \textcolor{green}{DCT}\} & \{KL, B, S, \textcolor{green}{LL}\} & \{KL, S, \textcolor{green}{KR}\}\\ 
        \textbf{RAPS} & \{C, PTS, \textcolor{green}{TAL}\} & \{IEC, C, PTS, L, P, , \textcolor{green}{DCT}\} & \{KL, B, S, \textcolor{green}{LL}\}  & \{B, KL, S, \textcolor{green}{KR}\}\\ 
        \textbf{Rand RAPS} & \{C, PTS, \textcolor{green}{TAL}\} & \{C, PTS, L, P, \textcolor{green}{DCT}\} & \{KL, S, \textcolor{green}{LL}\} & \{KL, S, \textcolor{green}{KR}\}\\ 
        \textbf{CPSN} & \{PTS, \textcolor{green}{TAL}\} & \{C, PTS, \textcolor{green}{DCT}\} & \{S, \textcolor{green}{LL}\} & \{S, \textcolor{green}{KR}\}\\ \toprule
    \end{tabular}
    \caption{A qualitative comparison of the conformal sets of CPSN vs. the tested baselines. Random samples were taken from the test fold, and the ground true classes are marked in \textcolor{green}{green}.}
    \label{tab:samples}
\end{table*}

\section{Experimental Results}
In this section we report the performance of the proposed CPSN method on several publicly available medical imaging classification tasks.

{\bf Datasets}. We used the \textbf{OrganAMNIST} and \textbf{TissuMNIST} datasets from the MedMNIST repository \cite{medmnistv1,medmnistv2}. OrganAMNIST is a dataset of Abdominal CT images with 11 body-organ classes. 
It is based on 3D CT images from Liver Tumor Segmentation Benchmark (LiTS) \cite{bilic2023liver}.  2D images were cropped from the center slices of the 3D bounding boxes in axial views and resized into $28 
 \times 28$. The dataset contains 58,850 images.
The TissueMNIST dataset
contains 236,386 human kidney cortex cells,  organized into 8 categories. Each gray-scale
image is $32 \times 32 \times 7$ pixels, where 7 denotes 7 slices. The  2D projections were obtained by taking the maximum pixel value along the axial-axis of each pixel, and were resized into $28 \times 28$ gray-scale images \cite{woloshuk2021situ}. 

To generate the input features (${x}$) and class probabilities (${p}$), we utilized trained CNNs published by MedMNIST authors \cite{medmnistv1,medmnistv2}. Each classifier is a ResNet-50 \cite{he2016deep} that has been fine-tuned over the train fold of one of the datasets. The logits/class probabilities as well as the pre-logits representation for each item were extracted from the validation and test sets. This results in 24,269 samples for the OrganAMNIST dataset and 70,920 samples for the TissueMNIST dataset.

{\bf Compared methods}. We compared the proposed CPSN method to the following prediction-set generation procedures: 1) The `naive' uncalibrated method \cite{angelopoulos2020uncertainty} (see a description in Section 2). 2)  The APS method \cite{romano2020classification} with and without  
a randomized procedure. 3)  The RAPS method 
\cite{angelopoulos2020uncertainty}
with and without a  randomized procedure.

{\bf Evaluation Measures}. The primary measure used for the evaluation of the prediction sets on
 a given test set are set size (average length of prediction sets) where a small value means high efficiency) and marginal coverage rate (fraction of testing examples for which the prediction sets contain the ground-truth labels). These two evaluation metrics can
be formally defined as:
$$ \textrm{size} = \frac{1}{n_{\textrm{test}}} \sum_i | C(x_i) |$$
$$ \textrm{coverage} = \frac{1}{n_{\text{test}}} \sum_i 
{\bf 1}(y_i \in C(x_i))$$
such that  $n_{\textrm{test}}$ is the size of the test set.

{\bf Implementation details}. Each dataset was divided into train/validation/test folds of $80\%/10\%/10\%$. {The first fold is used for training the regression network, the second for conformalization, and the third for evaluation (see Algorithm Box 1).} Temperature Scaling \cite{Guo2017} was used to calibrate the predicted class probabilities. APS was used to compute conformal scores for the training and validation sets. The regression network $q$ consists of a two-layered MLP with ReLU activation. As input, it receives a representation of 2048 dimensions and is optimized in order to predict the conformal score. Each network was trained for 100 epochs using AdamW optimizer \cite{loshchilov2017decoupled} with a learning rate of $5e-4$, weight decay of $1e-6$ and a batch size of 128. After convergence, the validation set is used for calculating the bias correction $\delta_{q(x)}$. The results are reported over the test fold. We repeat each experiment $10$ times using different train/validation/test split seeds.

Table~1 presents a summary of the experimental results. We report the mean and the standard deviation of 10 experiments for each setup for $\alpha=0.1$ and for $\alpha=0.05$. As can be seen, the coverage requirement was fulfilled by all the methods including `naive'. On average, CPSN produces small prediction sets with low variance for all the setups evaluated. The randomized versions of APS and RAPS yields better results than the deterministic versions. The RAPS algorithm became effective when the number of classes and the size of the prediction sets were both very large. As medical imaging classification tasks typically involve a moderate number of classes, APS and RAPS perform similarly. The bias correction values $\delta_1$ and $\delta_2$ in the case of TissueMNIST 
were $0.052\pm0.15$. and $ 0.134\pm0.17$.  This empirical evidence highlights the need to calculate the conformal bias correction separately for each of the two sets.


Table~2 presents a qualitative comparison of a few samples from the two datasets along with the prediction sets produced by CSPN and the baseline methods. It is evident that CSPN generats smaller prediction sets, even for the challenging samples, while maintaining a high coverage rate.

\section{Conclusions}
We presented a method that allows us to take any classifier network and generate predictive sets that are guaranteed to achieve a pre-specified error level, while maintaining a small average size. The main novelty of our approach, in comparison to previous CP methods, lies in the utilization of a sample-based threshold to form the prediction set which is computed by a network.
 In this study, we focused on the task of medical diagnostics based on medical images. 
Prediction sets are particularly valuable for medical doctors since they make it possible  to eliminate numerous possibilities and promptly refer the patient to the right specialists.
Our method, however, is versatile and can be applied to other critical tasks.

\section{Compliance with ethical standards}
\label{sec:ethics}

This research study was conducted retrospectively using
    human subject data made available in open access by (Source
    information). Ethical approval was not required as confirmed by
    the license attached with the open access data.
\balance
\bibliographystyle{IEEEbib}
\small
\bibliography{paper}




\end{document}